\documentclass[10pt]{article}
\date{}
\usepackage[left=2.2cm,right=2.2cm,top=2.5cm,bottom=2.5cm]{geometry}

\usepackage{newfloat}
\usepackage{graphicx}
\usepackage{newfloat}
\usepackage{graphics}
\graphicspath{ {./figures/} }
\usepackage{xcolor}
\usepackage{multirow}
\usepackage{numprint}
\usepackage{amsmath}
\usepackage{soul}
\usepackage{threeparttable}
\usepackage{booktabs}
\usepackage[font=bf,size=small]{caption}
\usepackage[font=normalfont]{subcaption}
\usepackage{hyperref}
\npthousandsep{,}\npthousandthpartsep{}
\usepackage{authblk}

\makeatletter
\g@addto@macro\normalsize{%
	\setlength\abovedisplayskip{1pt}
	\setlength\belowdisplayskip{1pt}
	\setlength\abovedisplayshortskip{1pt}
	\setlength\belowdisplayshortskip{1pt}
}
\makeatother

\newcommand{\NA}{---}

\title{Contextual Data Augmentation for Task-Oriented Dialog Systems}

\author{Dustin Axman \thanks{daxman@u.rochester.edu}}
\author{Avik Ray \thanks{avikray@amazon.com}}
\author{Shubham Garg \thanks{gargshu@amazon.com}}
\author{Jing Huang}
\affil{Amazon Alexa}

\begin{document}

\maketitle
 
\begin{abstract}
%Collection of annotated dialogs for training task-oriented dialog systems have been one of the key bottlenecks in improving current models. While dialog response generation has been widely studied on the agent side, it is not evident if similar generative models can be used to generate a large variety of, and often unexpected, user inputs that real dialog systems encounter in practice. Existing data augmentation techniques such as paraphrase generation techniques do not take the dialog context into consideration, and may not work well in real dialog settings. In this paper, we develop a novel dialog augmentation model that generates a user turn, conditioning on both past and future dialog context. Additionally, with a new context prompt design for language model, and output re-ranking, the augmented dialogs generated from our model can be directly used to train downstream task-oriented dialog systems. On common benchmark datasets MultiWoZ and SGD, we show that our dialog augmentation model generates high quality dialogs and improves dialog success rate by as much as $8\%$ over SOTA baseline. 

Collection of annotated dialogs for training task-oriented dialog systems have been one of the key bottlenecks in improving current models. While dialog response generation has been widely studied on the agent side, it is not evident if similar generative models can be used to generate a large variety of, and often unexpected, user inputs that real dialog systems encounter in practice. Existing data augmentation techniques such as paraphrase generation do not take the dialog context into consideration. In this paper, we develop a novel dialog augmentation model that generates a user turn, conditioning on full dialog context. Additionally, with a new prompt design for language model, and output re-ranking, the dialogs generated from our model can be directly used to train downstream dialog systems. On common benchmark datasets MultiWoZ and SGD, we show that our dialog augmentation model generates high quality dialogs and improves dialog success rate by as much as $8\%$ over baseline. 

% TODO: \jing{write how we use the pre-trained LM as novelty} 
\end{abstract}

\section{Introduction} \label{sec:intro}

% settings
Users of commercial voice assistants and chatbots (e.g. Alexa, Siri, Google assistant) are able to accomplish various tasks by interacting with them via natural language conversation. Task-oriented dialog models form the core technology behind these applications, which understands users' natural language utterances \cite{Hakkani-TurTCCG16,goo2018slot}, keeps track of the conversation \cite{MrksicSWTY17,ChenLWZT020}, performs requested tasks (e.g. API calls) \cite{WenMBY17,GaoWPLL18}, and generates appropriate meaningful response to the user \cite{WenGMSVY15,ZhangSGCBGGLD20}. 

% problem motivation and application, previous work, shortcomings, why is it challenging
Training neural task-oriented dialog models \cite{SimpleTOD:Hosseini-AslMWY20,SOLOIST:PengLLSLG21,UBAR:YangLQ21}, requires a large amount of annotated data, which is difficult to obtain for model developers. While crowd-sourcing and dialog simulation based on agent interplay \cite{shah2018building,Simulator:LinAEBZB20} addresses this issue to a certain extent, these are slow and don't provide sufficient coverage of different natural language (NL) user turn surface form variations. Recently, large pre-trained language models (e.g. GPT-2 \cite{GPT2:radford2019language}, T5 \cite{T5:RaffelSRLNMZLL20}) have been successfully used to generate fluent agent dialog responses, both with dialog context \cite{GuWWSY20,ZhangSGCBGGLD20,KaleR20} or without it \cite{KaleR20a,XuWKL20}. However, it is unclear if similar models can capture the large variation of user turn distribution in such task-oriented dialogs. Previous work on data augmentation for spoken language understanding has largely focused on generating paraphrases of user utterance, with a specific goal and set of entities \cite{HouLCL:18,ZhaoZY19,LinXZZZ21}. However, such utterances again fail to provide sufficient coverage of the large semantic space possible between dialog turns, and may not improve performance of downstream task-oriented dialog systems. As an example, in Table \ref{table:dialog1}, dialog $1$, the user says in the first turn $U_1=$ {\em ``please put me in touch with the local police, i was just robbed''}. A valid variation of this user turn which fits in the dialog context as generated by our model is $U_2 =$ {\em ``I was robbed and I am looking for help''}. Note that, $U_1$ and $U_2$ are not semantically equivalent paraphrases ($U_2$ doesn't explicitly request police).  

Therefore, in this paper, we propose a novel dialog augmentation model, using BART \cite{BART:LewisLGGMLSZ20}, which can generate variations of a user turn, when conditioned on past and future dialog turns. %and current belief state. 
Unlike dialog response generation, our model does not have the strict requirement of conveying a desired fixed action response, and can also leverage the future turns in the dialog. We show that using future context is indeed beneficial for the dialog augmentation task.
In addition, we propose a new NL context prompt design to delineate between user and system turns, %and dialog state slots/entities,
which better aligns with the language model pre-training task, and significantly improves quality of generated utterances and their positive impact on downstream tasks. %\textbf{(3)} We show that, using future context, is indeed beneficial for the dialog augmentation task. However, generating user turns with high semantic similarity to the original turn may not be very effective for this task.
On benchmark MutiWoZ \cite{Multiwoz21:EricGPSAGKGKH20} and SGD \cite{SGD:RastogiZSGK20} datasets, using dialogs generated from our augmentation model can significantly improve dialog success rate and goal accuracy compared to state-of-the-art baseline models.     

\section{Our Dialog Augmentation Model} \label{sec:model}

% overview and problem definition
In this section, we describe our dialog augmentation model. Let $D=\{U_i, B_i, S_i\}_{i=1}^n$ denote a task oriented dialog with $n$ turns, where $U_i$ denote the user request, $B_i$ represent the belief state, and $S_i$ denote the system response, for the $i$-th turn. We want to train a dialog augmentaion model $\mathcal{M}_d$ which can generate a user turn $U_t'$ for $t \in \left[n\right]$ given $D,$ such that $D' = \{U_1, B_1, \hdots, U_t', B_t, S_t, \hdots, B_n , S_n\}$ is a valid dialog. We refer $\{U_i,B_i,S_i\}_{i=1}^{t-1}$ as the {\bf past context} and $\{U_i,B_i,S_i\}_{i=t+1}^{n}$ as the {\bf future context}. 

%In the following, for clarity, we only describe how our model augments any $t$-th user turn. However, the same approach can be used to augment a system turn $S_t$ as well. In our experiments, we often train a single model which can augment both user and system turns. 

%\todo{Figure representing input/output of our model.} 

% pre-trained LM backbone
%{\bf Pre-trained language model backbone:} 
{\bf Base models:} Large pre-trained language models have been successfully used as backbone for common NLP tasks \cite{GPT2:radford2019language,BART:LewisLGGMLSZ20,T5:RaffelSRLNMZLL20}, modeling task-oriented dialog systems \cite{SimpleTOD:Hosseini-AslMWY20,SOLOIST:PengLLSLG21}, and dialog response generation \cite{KaleR20a,XuWKL20}. In this work we also use %based on transformer seq-to-seq architecture, 
BART as the base model for our dialog augmentation. BART is pre-trained with {\em text infilling} task where portions of the input text are masked using special mask tokens, and the model is tasked to re-generate these missing portions at the output. We leverage this task to augment dialog turns. Suppose we want to generate a variation of user turn $U_t$: we mask the $t$-th user turn, and construct the input sequence as $X=(U_1,S_1,\hdots,S_{t-1}, \left[\text{MASK}\right],S_{t+1},\hdots, U_n, S_n).$ The output is the user turn $Y=U_t.$ The base models are fine-tuned with $(X,Y)$ pairs by masking one turn for every dialog in the training set, and trained with cross-entropy loss. Note that, we do not use the belief states in this base model.  

% user/system turn tokens
{\bf Dis-entangling user/system turns:} Such a dialog augmentation model would struggle to learn the differences between user and system turn distribution, and often generate generic and uninformative turns such as {\em ``thank you''}, {\em ``you're welcome. enjoy!''}. To encourage the model to better learn the nuances of user and system turn distribution, we %also train a variant by 
add special user and system prompts (e.g. {\em user:/system:}) before every user $U_i$ and system $S_i$ turn respectively. We also add the user token before the mask tokens at the input, when we want to augment one user turn. Note that our model input design involves natural language prompts, as opposed to using special schema tokens (e.g. $\langle\text{user}\rangle$/$\langle\text{system}\rangle$) used in previous work \cite{KaleR20,LinXZZZ21,UBAR:YangLQ21}. This results in a better alignment of the input to BART's pre-training task and generates high quality of dialogs.

\begin{table}[t]
    \setlength\tabcolsep{3pt}
    \centering
    \small
    \caption{\label{table:multiwoz-ablation}Results of the our BART dialog augmentation model and its ablations on MultiWoZ 2.1 dataset. First two columns show extrinsic evaluation metrics, while the remaining columns present intrinsic evaluation metrics. }
    \begin{tabular}{p{3.3in} | l l | l p{0.5in} l}
    \toprule
        \textbf{Models} & \textbf{Inform} $\uparrow$ & \textbf{Success} $\uparrow$ & \textbf{BLEU} & \textbf{BERT Score} & \textbf{BLEURT} \\ 
        \midrule
        Soloist \cite{SOLOIST:PengLLSLG21} (no augmentation) & 0.873 & 0.733 & \NA & \NA & \NA \\ \hline
        T5 paraphrase augmentation & 0.903 & 0.722 & 0.592 & 0.956 & 0.307 \\ \hline
        Augmentation with past/future contexts & 0.930 & 0.794 & 0.188 & 0.890 & -0.547 \\
        %with Base BART Augmentation + $\langle user \rangle$/ $\langle system \rangle$ & 0.935 & 0.773 & 0.192 & 0.891 & -0.512 \\ 
        + ``user:/ system:'' & 0.910 & 0.772 & 0.196 & 0.890 & -0.518 \\ 
        %with Base BART Augmentation + "user/ system says" & 0.899 & 0.745 & 0.185 & 0.893 & -0.512 \\ 
        %with Base BART Augmentation + BS slots & 0.916 & 0.728 & 0.218 & 0.896 & -0.399 \\ 
        + re-rank & \textbf{0.936} & 0.751 & 0.188 & 0.890 & -0.547 \\
        %+ "user/system says" + BS Slots & 0.930 & 0.776 & 0.204 & 0.894 & -0.402 \\ \hline
        + ``user:/ system:'' + re-rank & 0.928 & \textbf{0.816} & 0.259 & 0.908 & -0.092 \\ 
        + ``user:/ system:'' + re-rank (no future context) &  0.917 & 0.740 & 0.171 & 0.887 & -0.431 \\ 
        + ``user:/ system:'' + re-rank + BS slots & 0.932 & 0.785 & 0.280 & 0.909 & -0.054 \\ 
        + ``user:/ system:'' + re-rank + BS slots (no future context) &  0.921 & 0.765 & 0.229 & 0.900 & -0.204  \\ 
        + BS slots & 0.916 & 0.728 & 0.218 & 0.896 & -0.399 \\ \bottomrule
    \end{tabular}
    
\end{table}

% output filtering/ranking
{\bf Output re-ranking:} % TODO: possibly elaborate on output filtering if we have room
Our post-generation re-ranking is done by generating 20 top augmentations (found through greedy search on a lattice with 25 beams). We compute the Bleurt score \cite{bleurt-sellam2020} between each generation and the true turn $U_t$ that we are currently augmenting. The highest-score augmentation is returned.

% discussion/comparison/ how is it different?
Note that, a key difference of our approach from previous dialog response generation work \cite{GuWWSY20,ZhangSGCBGGLD20,KaleR20} is that our model has access to both past/future dialog context; unlike response generation model which only has access to past turns. Additionally, response generation is typically a much more constrained problem due to its usage being generation of system responses in an online context, which conveys a specific intent/API response along with returned entities. In our dialog data augmentation problem, there is no need to restrict the output to be a strict paraphrase of the original user turn. Doing so harms the performance of downstream task-oriented dialog systems, as shown in Section \ref{sec:results}. Instead, we want the model to generate user turns that are rare and unseen in the training data, and that fit semantically within the provided past/future dialog context.

\section{Experiments} \label{sec:experiments}

In this section, we present intrinsic and extrinsic evaluation results of our proposed dialog augmentation model.

\begin{table}[ht]
    \setlength\tabcolsep{2pt}
    \centering
    \small
    \caption{\label{table:multiwoz-sampling} Extrinsic evaluation of our BART dialog augmentation model on MultiWoZ 2.1 dataset, in a low resource settings. We observe that augmenting data from our model helps downstream Soloist model achieve higher Inform and Success rates.}
    \begin{tabular}{p{2.75in} | l l | l l}
    \toprule
        \textbf{Models} & \multicolumn{2}{c|}{\textbf{$20\%$ data}} & \multicolumn{2}{c}{ \textbf{$50\%$ data}} \\ \hline
         & \textbf{Inform} $\uparrow$ & \textbf{Success} $\uparrow$ & \textbf{Inform} $\uparrow$ & \textbf{Success} $\uparrow$ \\
        \midrule
        Soloist \cite{SOLOIST:PengLLSLG21} (no augmentation) & 0.549 & 0.386 & 0.622 & 0.494 \\ \hline
        Augmentation w/ past/ future contexts + ``user:/ system:'' + re-rank  & 0.559 & 0.413 & 0.789 & 0.620 \\
        \bottomrule
    \end{tabular}
\end{table}

%\vspace{-4pt}

\subsection{Datasets} \label{sec:datasets}

%For our evaluation of dialog augmentation quality, we con
We experiment on common benchmark datasets MultiWoZ 2.1, and SGD for multi-domain task-oriented dialogs.

%{\bf MultiWoZ 2.1:} 
The {\bf MultiWoZ 2.1 dataset} \cite{Multiwoz21:EricGPSAGKGKH20}, is a consolidated and cleaned version of its earlier version \cite{MultiWOZ:BudzianowskiWTC18}. It is widely used as a benchmark for evaluation of task-oriented dialog models. This dataset contains task-oriented dialogs from multiple domains (e.g. Restaurant, Hotel, Attraction, Taxi, Train, Hospital,
Bus, and Police). The dataset contains \numprint{8438} training, \numprint{1000} dev, and \numprint{1000} test dialogs.  

The {\bf Schema Guided Dialog (SGD) dataset} \cite{SGD:RastogiZSGK20}, is a large task-oriented dialog benchmark dataset containing \numprint{22825} dialogs covering $16$ different domains (e.g. Flights, Hotels, Events, Services, Alarm etc.), and split into \numprint{16142} train, \numprint{2482} dev, and \numprint{4201} test dialogs. The dialogs are represented with a flexible and unified schema, which facilitates easier integration of new domain services via zero/few shot dialog state tracking.

\subsection{Metrics and setup} \label{sec:metrics}

Our evaluation is split into two groups, intrinsic and extrinsic. For {\bf intrinsic evaluation}, in MultiWoZ, we generate an augmented turn for every user turn of the test dialogs. We compute intrinsic evaluation metrics between augmented turn and ground truth user turn to gauge the augmentation quality. For the {\bf intrinsic metrics} we compute \textbf{BLEU} \cite{papineni-etal-2002-bleu}, \textbf{BertScore} \cite{bertscore-zhang2019}, \textbf{Bleurt} \cite{bleurt-sellam2020}. %We discuss these metrics in details in Appendix \ref{app:metrics}.

\begin{table}[t]
    \setlength\tabcolsep{4pt}
    \centering
    \small
    \caption{\label{table:sgd-ablation}Results of our BART dialog augmentation model and its ablations on SGD dataset using the SG--DST model \cite{SGD:RastogiZSGK20}.}
    \begin{tabular}{p{2.5in} | p{0.6in} p{0.6in} p{0.5in} p{0.5in} }
    \toprule
        \textbf{Models} & \textbf{Active Int Acc} $\uparrow$ & \textbf{Req Slot F1} $\uparrow$ & \textbf{Avg GA} $\uparrow$ & \textbf{Joint GA} $\uparrow$ \\ 
        \midrule
        SG--DST \cite{SGD:RastogiZSGK20} (no augmentation) & 0.870 & \textbf{0.968} & 0.559 & 0.241 \\ \hline
        Augmentation with past/future contexts & \textbf{0.902} & 0.965 & 0.569 & 0.249 \\ 
        %with Base BART Augmentation + $\langle user \rangle$/ $\langle system \rangle$ & 0.935 & 0.773 & 0.192 & 0.891 & -0.512 \\ 
        + ``user:/ system:'' & 0.901 & 0.965 & 0.560 & 0.250  \\ 
        %with Base BART Augmentation + "user/ system says" & 0.899 & 0.745 & 0.185 & 0.893 & -0.512 \\ 
        %with Base BART Augmentation + BS slots & 0.916 & 0.728 & 0.218 & 0.896 & -0.399 \\ 
        + re-rank &  0.898 &	0.965 & 0.572 & \textbf{0.257} \\ 
        %+ "user/system says" + BS Slots & 0.930 & 0.776 & 0.204 & 0.894 & -0.402 \\ \hline
        + ``user:/ system:'' + re-rank & 0.899 & 0.966 & \textbf{0.573} & 0.250 \\ 
        + ``user:/ system:'' + re-rank + BS slots & 0.901 & 0.966 & 0.570 & 0.244 \\  \bottomrule
    \end{tabular}
\end{table}

%Metric details are included in Appendix \ref{app:metrics}.

To evaluate use of the generated dialogs for helping downstream task-oriented dialog systems, we perform {\bf extrinsic evaluation} as follows. We augment $1$ randomly selected user turn in each of a fixed percentage $p$ of the training dialogs (we use $p=5\%$ in MultiWoZ, and $p=25\%$ for SGD\footnote{Since SGD is a larger dataset, we observed that it takes more augmentations to cause a significant improvement.}). These augmented dialogs (with the augmented turn replacing the original) are added into this training set. For MultiWoZ, we choose Soloist \cite{SOLOIST:PengLLSLG21} as the baseline, train it on this augmented training set and evaluate on the MultiWoZ test set using the most commonly used \textbf{Inform rate} and the \textbf{Success rate} metrics \cite{SOLOIST:PengLLSLG21}. In SGD, we select the dialog-state tracking (DST) baseline model Schema guided DST (SG--DST) introduced in \cite{SGD:RastogiZSGK20}. We evaluate the performance of SG--DST on the test split using the metrics {\bf Active intent accuracy}, {\bf Requested slots F1}, {\bf Average goal accuracy}, and {\bf Joint goal accuracy} as defined in \cite{SGD:RastogiZSGK20}. For each dataset we use the same extrinsic metrics as used in the original papers \cite{SOLOIST:PengLLSLG21,SGD:RastogiZSGK20} for easier comparison with corresponding baselines. On MultiWoZ, while models such as LAVA \cite{LAVALubisGHLMNG20} can achieve a better performance than Soloist, we consider it less suited for practical applications due to its complex multi-step RL based training.  
  
As a baseline {\bf paraphrase augmentation} model, we fine-tune a T5 model \cite{T5:RaffelSRLNMZLL20} using the paraphrase generation task on a set of paraphrases from MultiWoZ which were collected using the same method as in \cite{GaoZOY20}. Due to unavailability of such paraphrases for SGD dataset, we do not study this baseline for SGD.  

%baselines
%{\bf Baselines:} Our main baseline is the Soloist \cite{SOLOIST:PengLLSLG21} model trained without any data augmentation.  

% possibly remove if space is tough
\noindent{\bf Training details:}
We fine tune base BART model for 4 epochs on the task defined above, using batch size of $8$, and learning rate of $2 \times 10^{-5}.$ The encoder and embeddings are frozen during training. We use $4$ eval beams. For extrinsic evaluation, we train the Soloist model with default hyper-parameters using the original code \cite{SOLOISTCode}. The SGD--DST model was trained for $70$ epochs using the default hyper-parameters of the original implementation \cite{DSTCode}. We train both baseline models on the original training sets using the same hyper-parameters. All models were trained using a machine with single V100 GPU. Training BART dialog augmentation model requires about $3$ hours, training Soloist model takes approximately $7$ hours, and training SG--DST model takes about $26$ hours.

%We discuss more details in Appendix \ref{app:parameters}.

% \subsection{Baseline} \label{sec:baseline}

% For comparison, we used a 
%\subsection{Parameters} \label{sec:params}

%\subsection{Intrinsic evaluation}

%In this section, we present the results of intrinsic evaluation of our dialog augmentation model. For intrinsic evaluation, for each dataset, we consider the dialogs from the test set, and generate an augmented utterance for every turn of the dialog. We compare this generated utterance to the ground truth, and compute the various intrinsic evaluation metrics.

%\subsection{Extrinsic evaluation}

%We calculate our extrinsic metrics by using our method to augment 1 randomly selected User turn in each of 5\% of the training dialogs in MultiWoZ. These augmented dialogs (with the augmented turn replacing the original) are added into this training set. This increases the number of dialogs in the training set by 5\%.  The Soloist \cite{SOLOIST:PengLLSLG21} model is then trained on this set until the validation loss has decreased for 5 consecutive evaluations (conducted every 2000 steps) and evaluated on the MultiWoZ test set on the Extrinsic metrics Success Rate and Inform Rate.

%\dax{TODO: possibly add more on extrinsic if there's space}

%\subsection{Ablation study}

\subsection{Results and discussion} \label{sec:results}
% 5. BLEU and Bleurt are improved by BS slot prompting (intuitive)

% 1. Our method performs better on extrinsic evaluations than the baseline augmentation (not just better than the base no aug)
In Table~\ref{table:multiwoz-ablation} we present the results on MultiWoZ dataset. We observe that T5 paraphrase based augmentation achieves high intrinsic metrics since the paraphrases of the user turn are very similar to the original ground-truth user utterance. However, augmenting with paraphrases does not necessarily improve extrinsic metrics of downstream Soloist model. Our models achieve lower intrinsic metric compared to paraphrasing as expected, since they produce more variations of user turn, different from the ground-truth. However, the greater semantic coverage by our model leads to higher extrinsic metrics for downstream Soloist model. Our best augmentation model is the model with ``system:''/``user:'' prompting as well as a post augmentation re-ranking which significantly outperforms baseline Soloist model on all extrinsic metrics achieving an $8\%$ improvement in Success rate and $5\%$ improvement in Inform rate (row $6$ in Table~\ref{table:multiwoz-ablation}). This validates that augmentations generated from our model indeed help in improving downstream task-oriented dialog models. We also conclude that, traditional intrinsic metrics do not correlate well with extrinsic metrics for the dialog augmentation task. Since extrinsic metrics are more important for success of task-oriented dialog models, for the remaining section we mainly focus on these.

\begin{table}[ht]
    \setlength\tabcolsep{3pt}
    \centering
    \caption{\label{table:dialog1}Augmented dialog examples in MultiWoZ 2.1. The original turn with `*' around them (referred as GT) was augmented, and the augmented turn from our model (referred as AUG) is italicized.}
    \scriptsize
    \begin{tabular}{l p{4in}}
    \toprule
     & {\bf Dialog 1} \\ \midrule
    user (GT) &  *please put me in touch with the local police, i was just robbed.* \\
    user (AUG) &  {\em I was robbed and I am looking for help} \\
    system & i have information for the parkside police station, is this close to your location? \\
    user & i don't know, could you just give me the postcode and phone? \\
    system & the phone number is 01223358966 and the post code is cb11jg. \\
    user & thank you for your help \\
    system & you're welcome. the police will help you immediately. goodbye. \\
    \midrule 
    & {\bf Dialog 2} \\
    \midrule
    user & please see if you can find me a 3-star hotel on the east \\
 system & i'm sorry we do not have any east hotels that meet your criteria. would you like to try a different part of town? \\
user & yes, please look for any 3-star hotel -s in town that include free parking . east is still preferred. \\
system & none fits your criteria. can we change the area? \\
user & try something in the center of town. \\
system & the gonville hotel is an expensive, 3-star hotel in the centre of town. would you like me to book you a room? \\
user & yes please . it's for 2 people and 2 nights starting on thursday. \\
system & booking was successful. reference number is : 2v7ewg9e.  is there anything else i can help you with? \\
user (GT) & *how about a place to dine in the expensive price range serving indian food.* \\
user (AUG) & {\em I'm also looking for a place to dine that serves indian food} \\
system & i have many. what area would you prefer dining in? \\
user & i would like somewhere in the same area as the hotel. \\
system & i have 5 all located in the centre \\
user & then please pick 1 and give me their phone number band postcode. \\
system & saffron brasserie is located at hills road city centre, post code cb21la.  their phone number is 01223354679. can i help with anything else today? \\
user & that is everything . thank you for your help. \\
system & you are welcome. \\
    \bottomrule
\end{tabular}

\end{table}

Table \ref{table:sgd-ablation} presents the results on SGD with SG--DST (no augmentation) as the baseline. Although SGD is a much larger dataset compared to MultiWoZ, our BART augmentation model can still improve performance over baseline across several metrics ($3.2\%$ intent accuracy, $1.4\%$ average goal accuracy, and $1.6\%$ joint goal accuracy). We do not observe improvement in slot F1 score since we didn't perform any re-annotation of slots in the generated turn, which can result in some missing annotations. We also observe that ``system:''/ ``user:'' prompting is less beneficial than re-ranking in SGD. We hypothesize that in SGD it is easier for the model to differentiate user/ system distribution and thus making prompting less effective compared to MultiWoZ dataset.  

\noindent{\bf Results in low-resource settings:} 
%Annotated dialog datasets are difficult to collect, mainly for new dialog skill developers. -- Jing: these are more Alexa context that other people do not understand
Our dialog augmentation models can benefit the low resource dialog applications by boosting the performance of the dialog systems. We test this on the smaller MultiWoZ dataset, which can better emulate a low resource setting.  We further sample $20\%,$ and $50\%$ training dialogs from MultiWoZ training set, and augment additional $5\%$ of the sampled dialogs using our best performing model, BART + ``user: /system:'' prompts + re-rank. We then train the Soloist model on this augmented training data and evaluate the extrinsic metrics on the full test set. We compare the performance of this augmented model with baseline model trained only on the sampled training data ({\bf no augmentation}). From the results in Table~\ref{table:multiwoz-sampling} we observe that adding the augmented dialogs greatly improves both the Inform and Success rate of downstream Soloist model in low resource settings:
%by $1\%$ Inform rate and $2.7\%$ Success rate for the $20\%,$ data experiment, and 
by $16.7\%$ Inform rate and $12.6\%$ Success rate for the $50\%$ training data experiment. 
%-\todo{Discuss SGD low resource results.} 

\subsection{Ablation studies} 

We conduct several ablations designed to explore the impact of individual model components on our extrinsic and intrinsic metrics: the post augmentation re-ranking (``re-rank''), user/system prompts with (``user:/ system:''), removal of future context from the model input (``No future context''), as well as comparison with a variant that inserts belief state slots into the prompt (``BS slots''). The results are shown in Tables~\ref{table:multiwoz-ablation} and \ref{table:sgd-ablation}. We note that while increases in extrinsic metrics are objectively positive, intrinsic metrics are more open to interpretation. For example, having a high lexical/semantic similarity between an augmentation and original user turn does not always indicate the most useful augmentations from a downstream impact perspective. 

% 3. Most intrinsic metrics do not heavily correlate with extrinsic metrics.  We see that Bleurt correlates the most. Lower perplexity rarely indicates better extrinsic results (and why)

%We note that while increases in extrinsic metrics are objectively positive, intrinsic metrics are more open to interpretation. For example, having a high lexical similarity between an augmentation and input turn does not always indicate the most useful augmentations from a downstream impact perspective. Higher values of intrinsic metrics do not imply a better model for downstream augmentation.

% 2. Future context helps
\noindent{\bf Importance of future context:} To study if conditioning augmentations on future context indeed helps, we conduct an ablation on MultiWoZ where we remove the future context, and augment based on only past and current turns (no future context), which is the usual settings in response generation \cite{GuWWSY20,KaleR20a}. We observe that removing future context indeed harms downstream performance e.g. degrading Success rate by $7.6\%$ in our best model configuration. 
\noindent{\bf Effect of re-ranking:} We can see that re-ranking gives some of our best results in both SGD and MultiWoZ. In SGD, models with re-ranking achieve best goal accuracy, while in MultiWoZ extrinsic results are best when re-ranking is combined with user/system prompting. In all models except the base augmentation model, re-ranking improves scores on similarity metrics such as BERTScore, BLEU, and BLEURT. This is expected, because re-ranking based on Bleurt scores encourages selection of augmentations with greater semantic similarity to original user turn. 

%Current turn belief states overconstrain and don't improve metrics
\noindent{\bf Using belief state slots:} Using information from belief state $B_t,$ has been shown to be effective in generating relevant system dialog responses \cite{KaleR20,KaleR20a}. Our final ablation experiment ({\bf + BS slots}), studies the impact of adding entities/slots from belief state $B_t$ to the base model input $X$, just before the mask token. We convert the slots to their natural language template\footnote{\scriptsize For example, to augment user turn $U_t=$ {\em ``i need train reservations from norwich to cambridge''} containing entities \{``norwich'', ``cambridge''\}, we include natural language template phrase {\em ``train departing norwich, train destination cambridge''} before the mask token in the input $X$.} similar to \cite{ZhaoZY19,KaleR20}. This encourages the model generate an augmented turn $U_t'$ containing the same set of entities. This is also validated by a consistent increase in intrinsic semantic similarity metrics when BS slots are added to the input. Although proven to be effective in response generation, for both MultiWoZ and SGD, it does not offer improvement over our best model in extrinsic metrics. We hypothesize that this is because these slots over-constrain the augmentation to generate close paraphrases of the original user turn.

\subsection{Example augmented dialogs}

In the Table \ref{table:dialog1} we present some example dialogs generated by our augmentation model in MultiWoZ dataset. The turn with `*' around them was augmented with our model (also referred as ground-truth GT). The augmented dialog generated by our model is italicized (also referred as AUG). In {\bf Dialog 1} we can see that our augmentation of the first turn is very similar to the original even though it was generated without seeing the original turn as input. In {\bf Dialog 2}, augmentation preserves the user intent and entity ``indian'', but drops the term ``expensive''. In future work, we want to research more into effective ways of leveraging this current turn entity information. In both the examples, our model generates a new user turn variation which fits in the dialog context, although not being a strict paraphrase of the original user turn. 

\section{Conclusion} \label{sec:conclusion}
We develop a novel dialog augmentation model, which can generate new user turn variations given both past and future dialog context, and current dialog state. We carefully design natural language prompts for pre-trained language models. Together with a re-ranking model, our data augmentation approach generates high quality dialogs that can augment existing  dialog datasets. We further show that the augmentation data from our model greatly improve dialog completion and success rates of SOTA task-oriented dialog systems. Using ablation study, we also highlight an important tradeoff: generating accurate paraphrases of user turns does not necessarily improve downstream task-oriented dialog systems. Instead, generating more variations of user inputs that fits the given context, would result in better performance.

\bibliographystyle{IEEEtran}
\bibliography{dialog_short}

\begin{thebibliography}{10}
\providecommand{\url}[1]{#1}
\csname url@samestyle\endcsname
\providecommand{\newblock}{\relax}
\providecommand{\bibinfo}[2]{#2}
\providecommand{\BIBentrySTDinterwordspacing}{\spaceskip=0pt\relax}
\providecommand{\BIBentryALTinterwordstretchfactor}{4}
\providecommand{\BIBentryALTinterwordspacing}{\spaceskip=\fontdimen2\font plus
\BIBentryALTinterwordstretchfactor\fontdimen3\font minus \fontdimen4\font\relax}
\providecommand{\BIBforeignlanguage}[2]{{%
\expandafter\ifx\csname l@#1\endcsname\relax
\typeout{** WARNING: IEEEtran.bst: No hyphenation pattern has been}%
\typeout{** loaded for the language `#1'. Using the pattern for}%
\typeout{** the default language instead.}%
\else
\language=\csname l@#1\endcsname
\fi
#2}}
\providecommand{\BIBdecl}{\relax}
\BIBdecl

\bibitem{Hakkani-TurTCCG16}
D.~Hakkani{-}T{\"{u}}r, G.~T{\"{u}}r, A.~Celikyilmaz \emph{et~al.}, ``Multi-domain joint semantic frame parsing using bi-directional {RNN-LSTM},'' in \emph{Interspeech}, 2016.

\bibitem{goo2018slot}
C.-W. Goo, G.~Gao, Y.-K. Hsu \emph{et~al.}, ``Slot-gated modeling for joint slot filling and intent prediction,'' in \emph{NAACL-HLT}, 2018.

\bibitem{MrksicSWTY17}
N.~Mrksic, D.~{\'{O}}. S{\'{e}}aghdha, T.~Wen \emph{et~al.}, ``Neural belief tracker: Data-driven dialogue state tracking,'' in \emph{ACL 2017}, 2017.

\bibitem{ChenLWZT020}
L.~Chen, B.~Lv, C.~Wang \emph{et~al.}, ``Schema-guided multi-domain dialogue state tracking with graph attention neural networks,'' in \emph{AAAI}, 2020.

\bibitem{WenMBY17}
T.~Wen, Y.~Miao, P.~Blunsom \emph{et~al.}, ``Latent intention dialogue models,'' in \emph{{ICML}}, 2017.

\bibitem{GaoWPLL18}
B.~Peng, X.~Li, J.~Gao, J.~Liu, and K.~Wong, ``Deep dyna-q: Integrating planning for task-completion dialogue policy learning,'' in \emph{{ACL}}, 2018.

\bibitem{WenGMSVY15}
T.~Wen, M.~Gasic, N.~Mrksic, P.~Su, D.~Vandyke, and S.~J. Young, ``Semantically conditioned lstm-based natural language generation for spoken dialogue systems,'' in \emph{{EMNLP}}, 2015.

\bibitem{ZhangSGCBGGLD20}
Y.~Zhang, S.~Sun, M.~Galley \emph{et~al.}, ``{DIALOGPT} : Large-scale generative pre-training for conversational response generation,'' in \emph{{ACL}}, 2020.

\bibitem{SimpleTOD:Hosseini-AslMWY20}
E.~Hosseini{-}Asl, B.~McCann, C.~Wu \emph{et~al.}, ``A simple language model for task-oriented dialogue,'' in \emph{NeurIPS}, 2020.

\bibitem{SOLOIST:PengLLSLG21}
B.~Peng, C.~Li, J.~Li \emph{et~al.}, ``{SOLOIST:} building task bots at scale with transfer learning and machine teaching,'' \emph{TACL}, vol.~9, 2021.

\bibitem{UBAR:YangLQ21}
Y.~Yang, Y.~Li, and X.~Quan, ``{UBAR:} towards fully end-to-end task-oriented dialog system with {GPT-2},'' in \emph{{AAAI}}, 2021.

\bibitem{shah2018building}
P.~Shah, D.~Hakkani-T{\"u}r, G.~T{\"u}r \emph{et~al.}, ``Building a conversational agent overnight with dialogue self-play,'' \emph{arXiv arXiv:1801.04871}, 2018.

\bibitem{Simulator:LinAEBZB20}
C.-W. Lin, V.~Auvray, D.~Elkind \emph{et~al.}, ``Dialog simulation with realistic variations for training goal-oriented conversational systems,'' \emph{arXiv preprint arXiv:2011.08243}, 2020.

\bibitem{GPT2:radford2019language}
A.~Radford, J.~Wu, R.~Child \emph{et~al.}, ``Language models are unsupervised multitask learners,'' \emph{OpenAI blog}, 2019.

\bibitem{T5:RaffelSRLNMZLL20}
C.~Raffel, N.~Shazeer, A.~Roberts \emph{et~al.}, ``Exploring the limits of transfer learning with a unified text-to-text transformer,'' \emph{JMLR}, vol.~21, 2020.

\bibitem{GuWWSY20}
J.~Gu, Q.~Wu, C.~Wu, W.~Shi, and Z.~Yu, ``{PRAL:} {A} tailored pre-training model for task-oriented dialog generation,'' in \emph{{ACL/IJCNLP}}, 2021.

\bibitem{KaleR20}
M.~Kale and A.~Rastogi, ``Template guided text generation for task-oriented dialogue,'' in \emph{{EMNLP}}, 2020.

\bibitem{KaleR20a}
------, ``Text-to-text pre-training for data-to-text tasks,'' in \emph{{INLG}}, 2020.

\bibitem{XuWKL20}
X.~Xu, G.~Wang, Y.~Kim, and S.~Lee, ``Augnlg: Few-shot natural language generation using self-trained data augmentation,'' in \emph{{ACL/IJCNLP}}, 2021.

\bibitem{HouLCL:18}
Y.~Hou, Y.~Liu, W.~Che, and T.~Liu, ``Sequence-to-sequence data augmentation for dialogue language understanding,'' in \emph{Proc. of {COLING}}, 2018.

\bibitem{ZhaoZY19}
Z.~Zhao, S.~Zhu, and K.~Yu, ``Data augmentation with atomic templates for spoken language understanding,'' in \emph{Proc. of {EMNLP-IJCNLP}}, 2019.

\bibitem{LinXZZZ21}
H.~Lin, L.~Xiang, Y.~Zhou, J.~Zhang, and C.~Zong, ``Augmenting slot values and contexts for spoken language understanding with pretrained models,'' in \emph{Interspeech}, 2021.

\bibitem{BART:LewisLGGMLSZ20}
M.~Lewis, Y.~Liu, N.~Goyal \emph{et~al.}, ``{BART:} denoising sequence-to-sequence pre-training for natural language generation, translation, and comprehension,'' in \emph{{ACL}}, 2020.

\bibitem{Multiwoz21:EricGPSAGKGKH20}
M.~Eric, R.~Goel \emph{et~al.}, ``Multiwoz 2.1: {A} consolidated multi-domain dialogue dataset with state corrections and state tracking baselines,'' in \emph{{LREC}}, 2020.

\bibitem{SGD:RastogiZSGK20}
A.~Rastogi, X.~Zang, S.~Sunkara \emph{et~al.}, ``Towards scalable multi-domain conversational agents: The schema-guided dialogue dataset,'' in \emph{{AAAI}}, 2020.

\bibitem{bleurt-sellam2020}
T.~Sellam, D.~Das, and A.~P. Parikh, ``Bleurt: Learning robust metrics for text generation,'' \emph{arXiv preprint arXiv:2004.04696}, 2020.

\bibitem{MultiWOZ:BudzianowskiWTC18}
P.~Budzianowski, T.~Wen \emph{et~al.}, ``Multiwoz - {A} large-scale multi-domain wizard-of-oz dataset for task-oriented dialogue modelling,'' in \emph{{EMNLP}}, 2018.

\bibitem{papineni-etal-2002-bleu}
K.~Papineni, S.~Roukos, T.~Ward, and W.-J. Zhu, ``{B}leu: a method for automatic evaluation of machine translation,'' in \emph{{ACL}}, 2022.

\bibitem{bertscore-zhang2019}
T.~Zhang, V.~Kishore, F.~Wu, K.~Q. Weinberger, and Y.~Artzi, ``Bertscore: Evaluating text generation with {BERT},'' in \emph{{ICLR}}, 2020.

\bibitem{LAVALubisGHLMNG20}
N.~Lubis, C.~Geishauser, M.~Heck, H.~Lin, M.~Moresi, C.~van Niekerk, and M.~Gasic, ``{LAVA:} latent action spaces via variational auto-encoding for dialogue policy optimization,'' in \emph{{COLING}}, 2020, pp. 465--479.

\bibitem{GaoZOY20}
S.~Gao, Y.~Zhang, Z.~Ou, and Z.~Yu, ``Paraphrase augmented task-oriented dialog generation,'' in \emph{{ACL}}, 2020.

\bibitem{SOLOISTCode}
{Baolin Peng and Chunyuan Li and Jinchao Li and Shahin Shayandeh and Lars Liden and Jianfeng Gao}, ``Soloist,'' \url{https://github.com/pengbaolin/soloisthttps://github.com/pengbaolin/soloist}, 2021.

\bibitem{DSTCode}
{Rastogi et al.}, ``Schema guided dst,'' \url{https://github.com/google-research/google-research/tree/master/schema_guided_dst}, 2020.

\end{thebibliography}

\end{document}